\pdfoutput=1 

\documentclass{article}
\usepackage{spconf,amsmath,graphicx}
\usepackage[misc]{ifsym} 
\usepackage{booktabs}
\usepackage{amsmath,amsthm,amssymb,mathrsfs,enumitem}
\usepackage{multirow}
\usepackage{microtype, dsfont}
\usepackage{xspace}
\usepackage{cite}
\usepackage{algorithmic}
\usepackage{textcomp}
\usepackage{xcolor}
\usepackage{CJK}

\def\BibTeX{{\rm B\kern-.05em{\sc i\kern-.025em b}\kern-.08em
    T\kern-.1667em\lower.7ex\hbox{E}\kern-.125emX}}
\begin{document}

\title{Position-based Contributive Embeddings for Aspect-Based Sentiment Analysis}

\name{Zijian Zhang$^{1 2}$ \qquad Chenxi Zhang$^{1}$ \qquad Jiangfeng Li$^{1 2}$*\thanks{* The corresponding author} \qquad Qinpei Zhao$^{1}$}
  
\address{$^{1}$ School of software, Tongji University, Shanghai, China \\
         $^{2}$ Key Laboratory of Blockchain and Cyberspace Governance of Zhejiang Province, Hangzhou, China}

\begin{CJK}{UTF8}{gbsn}
\maketitle

\begin{abstract}
Aspect-based sentiment analysis (ABSA), exploring sentiment polarity of aspect-given sentence, is a fine-grained task in the field of nature language processing. Previously researches typically tend to predict polarity based on the meaning of aspect and opinions.  However, those approaches mainly focus on considering relations implicitly at the word level, ignore the historical impact of other positional words when the aspect appears in a certain position. Therefore, we propose a Position-based Contributive Embeddings (PosCE) to highlight the historical reference to special position aspect. Contribution of each positional words to the polarity is similar to the process of fairly distributing gains to several actors working in coalition (game theory). Therefore, we quote from the method of Shapley Value and finally gain PosCE to enhance the aspect-based representation for ABSA task. Furthermore, the PosCE can also be used for improving performances on multimodal ABSA task. Extensive experiments on both text and text-audio level using SemEval dataset show that the mainstream models advance performance in accuracy and F1 (increase 2.82\% and 4.21\% on average respectively) by applying our PosCE.
\end{abstract}
\begin{keywords}
Shapley value, PosCE, ABSA
\end{keywords}

\section{Introduction}
Constructing Aspect-Based Sentiment Analysis (ABSA) methods is one of the most challenging tasks in researches of human language processing. It aims to judge sentiment polarity of an aspect-given sentence rather than the whole sentence. Language experts show that most of errors are caused by ignoring the aspect information, other than the choice of downstream models \cite{b1} (like RNN, BERT, etc.). Thus, capturing aspect signals and generating aspect-related representation for classification model is the kernel of ABSA. 
 
Existing works for ABSA have adopted a two-step approach \cite{b2}: extract aspect terms and opinion terms for classification respectively. The traditional methods use low-dimensional and dense vectors to implicitly represent semantic feature, but performs poorly \cite{b3}. Recently, the success of deep learning has inspired various neural network architectures. Some researchers adjust the structure of the model, such as Convolutional Neural Network (CNN) \cite{b4}, Recursive Neural Networks (RNN) \cite{b5} and Pre-trained Language Models (PLMs) \cite{b6}. However, these model pay less attention to the syntactic relations among aspect terms and opinion terms. Hence, feature-based methods are proposed to gain sentence representation with aspect signals, which the most influential mechanism is attention. Nevertheless, the distance-based attention can only capture low-sensitive of distance aware, which is a disaster in NLP tasks\footnote{Low-sensitive distance aware generally refers to the words order through sentence can not be recognized and modeled in NLP field.}. Although positional embeddings is proposed to partly solve this problem, it can only distinguish position difference rather than the essential difference of position feature \cite{b7}. Consequently, a historical records based method should be used for ABSA task. It can learn a pattern: when an aspect item appears in a specific position in a sentence, what should be the contribution of words in other positions to polarity.

In this paper, a novel Position-guided Contributive Embeddings (PosCE) is proposed to cultivate aspect-based contribution for sentence representation as well as improve the performance of mainstream models. The significant application of PosCE is based on the assumption that aspect terms in the specific location are affected by other regular positional words, especially the opinion terms. The major difference between PosCE and other feature-based methods is that PosCE displays the position of aspect terms and counts the contribution of other places from the historical data based on Shapley Value \cite{b8}. Basicly, a sentence splits into left context, aspect term and right context. Next, the phrases are converted into vectors by token embeddings. After that, considering contextual combinations with aspect term as each combination is calculated. For example, A\textbf{\textcolor{red}{\emph{B}}}C is a word-level split sentence of ABSA task. Then the combinations are \{A\textbf{\textcolor{red}{\emph{B}}},  \textbf{\textcolor{red}{\emph{B}}}C,  A\textbf{\textcolor{red}{\emph{B}}}C\}. Considering all combinations, the average pooling of marginal contribution will be taken as the outputs of PosCE, which can reflect the contribution of other positions to the final predictions. Recall that the position-guided contributive embeddings is decided by the certain position of aspect.  In conclusion, the contributions of our paper are as follows:
 
\begin{itemize}[topsep=1pt]
\setlength{\itemsep}{0pt}
\setlength{\parsep}{0pt}
\setlength{\parskip}{0pt}
\item Position-guided Contributive Embeddings is proposed to capture contextual regulation for feature enhancment. 
\item A flexible training method for PosCE, which can update once or more times in mainstream models.
\item Extensive experminets on SemEval dataset show the effectiveness of PosCE.
\end{itemize}

The follows of paper will describes the problem we solved and shows the proposed methods in details. Then, the performance of PosCE and the conclusion will be proposed.


\section{Problem Definition}
\subsection{Illustration of Aspect-Based Sentiment Analysis}
\begin{figure}[t]
\centerline{\includegraphics[scale=0.35]{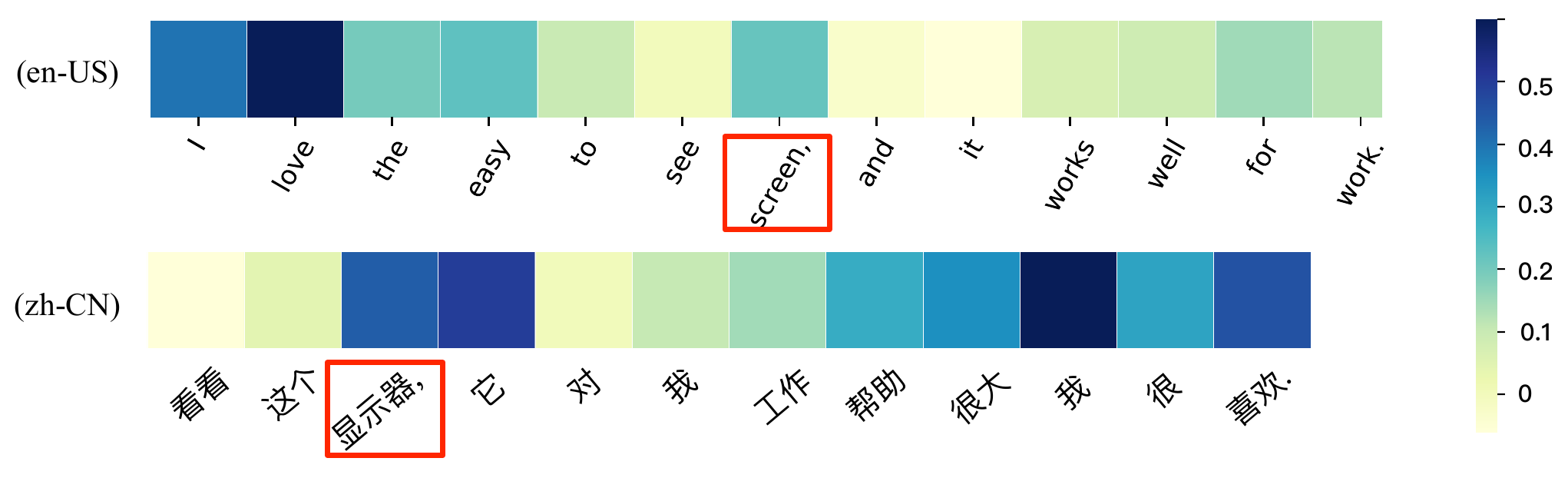}}
\caption{Example of a sentece with its PosCE.}
\label{fig}
\end{figure}

The ABSA task aims to predict the sentiment polarity of a sentence towards aspect term. An example of this task is given here\footnote{\emph{“\underline{I love} the easy to see \textcolor{red}{\textbf{screen}}, which \underline{works well for work}.}” is an example in SemEval2014 datasets. The \textcolor{red}{\emph{\textbf{screen}}} is the aspect term, and the whole sentence shows positive sentiment towards aspect through the \underline{underline words}.}. Figure 1 shows an example of Position-guided contributive embeddings, and the contribution of different locational words indicated by color depth. The \textcolor{red}{\textbf{\emph{screen}}} (or \textcolor{red}{\textbf{\emph{显示器}}}) is uniquely in the $7^{th}$ (or $3^{th}$ in Chinese case) position and the contexts affect the polarity regularly. Utilizing PosCE to solve classification is regarded as the only difference between traditional sentient analysis and ABSA in this paper.

\subsection{Formal Definition}
Let $S=\{w_{i=1,...,t,...,L}\}$ to be a sentence with $L$ words and the aspect term $w_t$ at $t^{th}$ position. The tokens embeddings $F^{tok} \in \mathbb{R}^{L \times k}$ together with other two adding features: positional embeddings $F^{pos}$ and contributive embeddings $F^{PosCE}_t$ are input into the classifier, which can be defined as: $F = F^{tok} \oplus F^{pos} \oplus F^{PosCE}_t$ (in Figure 2). The objective of ABSA is to assign sentiment polarity $P \in \mathbb{R}^3$, where $P = \{positive, neutral,negative \}$. 
\begin{figure}[b]
\centerline{\includegraphics[scale=0.25]{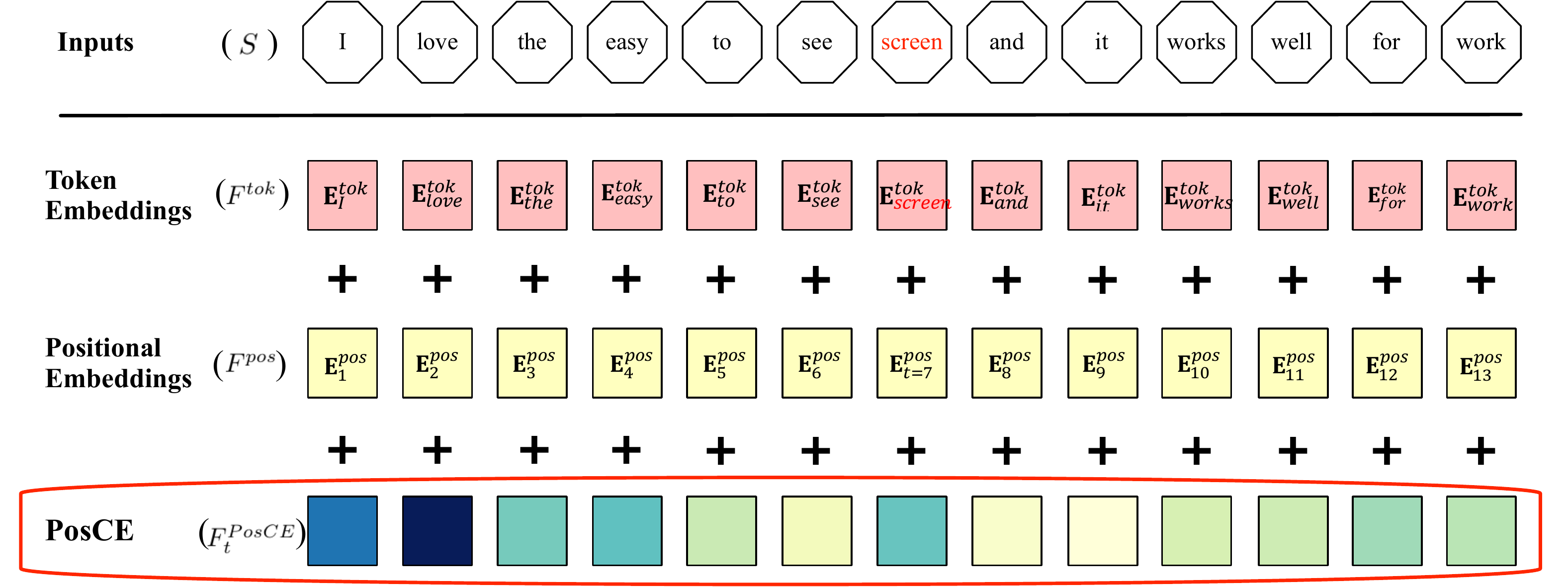}}
\caption{Input Representation.}
\label{fig}
\end{figure}

\subsection{Problem Statement}
Under the illustration and formal descirption, the scheme can be divided into two basic step, which are desribed as follows.
\begin{itemize}
\setlength{\itemsep}{0pt}
\setlength{\parsep}{0pt}
\setlength{\parskip}{0pt}
\item[1)] \textbf{ABSA task with PosCE}. Given sentence $S$ with aspect term in position $t$ to explore polarity, which special in:
\begin{itemize}[topsep=1pt]
\setlength{\itemsep}{0pt}
\setlength{\parsep}{0pt}
\setlength{\parskip}{0pt}
\item[a.] Considering the representation $F$ with $F^{PosCE}_t$;
\item[b.] PosCE can be updated during training;
\item[c.] For any phrases combinations have its corresponding predicted value $P \in \mathbb{R}^3$.
\end{itemize}
\item[2)] \textbf{The solution of PosCE}.This paper model the historical references by embedding mechanism, that is how context is regularly related to aspect $w_t$. The prediction of phrases $P = \{S-\cup_i \{w_i\}|i \neq t\}$ is calculated as the marginal contribution. Then the PosCE $(v^t_1,...,v^t_L), v^t_i = \frac{1}{L}\sum_{i \neq t} P_i$ is obtained by using the method of Shapley value.
\end{itemize}


\section{Methodology}
In this section, we describe our proposed embeddings method and the usage in mainstream models. Both model architecture and the PosCE procedure are depicted in Figure 3.

\subsection{Model Structure}
\begin{figure*}[t]
\centerline{\includegraphics[scale=0.35]{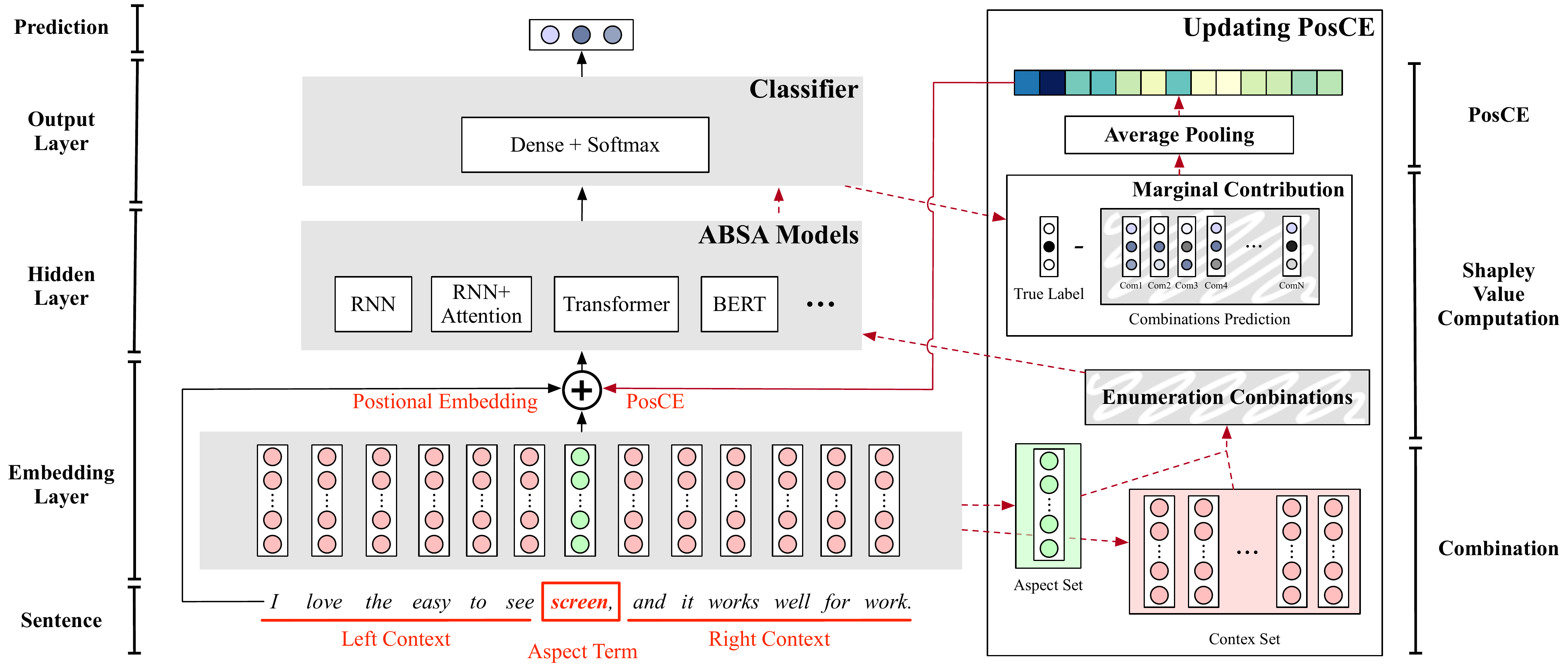}}
\caption{Overview of PosCE. Solving ABSA with PosCE are in the left and the method to update PosCE are in the right.}
\label{fig}
\end{figure*}

The ABSA model consists of special \textbf{embedding layer} (adding PosCE) and \textbf{classification modules}. The token embeddings $F^{tok}$ together with $F^{pos}$ and $F^{PosCE}_t$ are the elemenst of embedding layer, which are feed forward to the following modules. Recall that the value of $F^{PosCE}_t$ varies according to the position $t$ (aspect term). The hidden layer is simplified to $M \in \mathbb{R}^{k \times h}$, where $h$ is hidden dimension. In this way, the middle matrix $H$ of the inputs is obtained by:
\begin{equation}
    H = \textbf{\rm tanh}(F \dot M + b)
\end{equation}

Then, the final probability prediction distribution can be gained by the output layer.
\begin{equation}
    P =Softmax(\textbf{Dense}(H))
\end{equation} where $\textbf{Dense}$ denotes complex networks, including pooling for sentence vectors and fully connection for classification. 

For each sentence input into ABSA model, a probability distribution can be obtained. The value $P_{true}$ corresponding to the certain position of ground truth can be understood as the prediction result of the input. The difference between combination \{A\textbf{\textcolor{red}{\emph{B}}}C\} and \{A\textbf{\textcolor{red}{\emph{B}}}\} can represent the marginal contribution of word C. When all combinations are counted, the contribution of $3^{th}$ (word C) position can be obtained by average. This situation is extremely similar to the n-person cooperation in game theory to seek everyone's contribution to the group. Therefore, this paper draws lessons from the Shapley Value to obtain PosCE.

\subsection{Position-Guided Contributive Distribution unit}
When aspect term is given, the sentiment of sentence towards this term is affected by other positional words regularly. The PosCE can learn this pattern, which captures information from aspect term and its surrounding contexts simultaneously.

The method of capturing PosCE is shown on the right of Figure 3. Firstly, enumeration combinations generated by aspect set $\{w_t\}$ and context set $C$, where $|\{w_t\}|+|C|=L$. There are $\Gamma = s^{|C|}-1$ phrases combinations in total. Using the computation method of Shapley value, each marginal contribution of combinations can be calculated by formula (2). In other words, it's equal to the $P_{true}$ mentioned before. Shapley value for special aspect words or position is defined as:
\begin{equation}
    \Phi_t(N, w_t) = \frac{(|N|-|C^*|)|C^*|!}{|N|!}[P(C^* \cup {w_t})-P(C^*)]
\end{equation} where $C^*$ is a subset of context sets $C$ and $w_t$ is an element of $N$. Briefly, the Shapley value of given $w_t$ is the weighted sum of the contribution in each subset of $C$. Meanwhile, function $\Phi_t(N, w_t)$ and $P(C^* \cup {w_t})-P(C^*)$ can be abbreviated as $\Phi_t(N)$ and $P_t(C^*)$. The contribution of $w_t$ with respect to set $C^*$ is computed by $P(C^* \cup {w_t})-P(C^*)$.

The Shapley value is considered as a uniquely fair way for distributing the true label $P(N)$ into $(\Phi_1(N),...,\Phi_L(N))$ for $L$ words, since it satisfies the following characteristics \cite{b9}:
\begin{itemize}[topsep=1pt]
\setlength{\itemsep}{2pt}
\setlength{\parsep}{0pt}
\setlength{\parskip}{0pt}
\item Efficiency: $\sum_{1 \leq t \leq L} \Phi_t(N, w_t) = P_t(N)$.
\item Symmetry: For words $w_i$ and $w_j$, if $\forall{S^*}\subseteq N-\{w_i, w_j\} \& P_i(S*) = P_j(S^*)$, then $\Phi_i(N)=\Phi_j(N)$.
\item Dummy: If $\forall{S^*} \subseteq N-\{w_t\}$ and $P_t(S^*)=P(S^*)$, then $P(w_t)=0$. 
\item Additivity: For any pair of combinations $<S, w_i>$ and $<S, w_j>: \Phi_{i+j}(S)=\Phi_i(S)+\Phi_j(S)$ where $S^* \subseteq S$, then $\forall{S^*}, \phi_{i+j}(S^*)\phi_{i}(S^*)+\phi_{j}(S^*, w_j)$.
\end{itemize}

In order to gain PosCE towards special position, an average pooling should be used. Then the whole PosCE $P^{PosCE}_t \in \mathbb{R}^L$ towards $t^{th}$ position can be formulated as:
\begin{equation}
    P^{PosCE}_t = Softmax(\frac{1}{\Gamma} \sum_{S^* \in S} \Phi_t(S^*, w_t))
\end{equation}

\begin{table*}[ht]
\footnotesize
\caption{Experiment results for text-level and text-audio ABSA task. Accuracy and F1 is reported, where \textcolor{blue}{$\uparrow$} indictes an increase of less than $3\%$, \textcolor{orange}{$\uparrow$} indicates between $3\%$ and $5\%$ and \textcolor{red}{$\uparrow$} indicates more than $5\%$.}
\centering
\renewcommand\arraystretch{0.9}
\begin{tabular}{clcccccc|cccccc}
\toprule
& \multicolumn{1}{c}{}   & \multicolumn{6}{c|}{without PosCE}  & \multicolumn{6}{c}{with PosCE}     \\ \hline
& \multicolumn{1}{c}{\multirow{2}{*}{Model}} & \multicolumn{2}{c}{Laptop} & \multicolumn{2}{c}{Restaurant} & \multicolumn{2}{c|}{Twitter} & \multicolumn{2}{c}{Laptop} & \multicolumn{2}{c}{Restaurant} & \multicolumn{2}{c}{Twitter} \\
& \multicolumn{1}{c}{} & Acc  & F1  & Acc  & F1  & Acc  & F1  & Acc  & F1   & Acc   & F1  & Acc  & F1  \\ \hline
\multirow{6}{*}{\begin{tabular}[c]{@{}c@{}}\textbf{Single}\\ \textbf{Modal}\end{tabular}} & SVM    & 69.52 & 0.687 & 77.26 & 0.762 & 62.17  & 0.612  & -   & -   & -    & -    & -    & -            \\
& LSTMs(${\rm RNN}$)  & 68.70  & 0.601  & 78.29  & 0.771  & 67.89  & 0.641  & 70.47 \textcolor{blue}{$\uparrow$} & 0.681 \textcolor{red}{$\uparrow$} & 78.12  & 0.776 \textcolor{blue}{$\uparrow$}  & 70.27 \textcolor{orange}{$\uparrow$} & 0.689 \textcolor{blue}{$\uparrow$}\\
& AOA(${\rm RNN}$)        & 74.91  & 0.716  & 78.61  & 0.741  & 69.28  & 0.675  & 75.05 \textcolor{blue}{$\uparrow$} & 0.717 \textcolor{blue}{$\uparrow$} & 79.81 \textcolor{blue}{$\uparrow$}  & 0.801 \textcolor{orange}{$\uparrow$}  & 72.17 \textcolor{orange}{$\uparrow$} & 0.701 \textcolor{orange}{$\uparrow$} \\
& TNet(${\rm Trans}$)       & 76.54  & 0.715  & 80.27  & 0.796  & 71.06  & 0.691  & 79.14 \textcolor{orange}{$\uparrow$} & 0.768 \textcolor{red}{$\uparrow$} & 81.72 \textcolor{blue}{$\uparrow$}  & 0.810 \textcolor{blue}{$\uparrow$}  & 75.23 \textcolor{red}{$\uparrow$} & 0.745 \textcolor{red}{$\uparrow$}\\
& BERT-SPC(${\rm Trans}$)   & 79.47  & 0.765  & 82.17  & 0.804  & 74.71  & 0.715  & 81.49 \textcolor{orange}{$\uparrow$} & 0.795 \textcolor{orange}{$\uparrow$} & 84.25 \textcolor{orange}{$\uparrow$}  & 0.836 \textcolor{orange}{$\uparrow$}  & 77.19 \textcolor{orange}{$\uparrow$} & 0.741 \textcolor{orange}{$\uparrow$}\\ 
\bottomrule
\multirow{3}{*}{\begin{tabular}[c]{@{}c@{}}\textbf{Multi}\\ \textbf{Modal}\end{tabular}}  & MDRE    & 67.48  & 0.662  & 73.96  & 0.717  & 62.77 & 0.607  & 67.74 \textcolor{blue}{$\uparrow$}  & 0.670 \textcolor{blue}{$\uparrow$}  & 73.42  & 0.720  \textcolor{blue}{$\uparrow$}  & 64.39  \textcolor{blue}{$\uparrow$}  & 0.619 \textcolor{blue}{$\uparrow$}       \\
& MCNN-LSTM  & 65.61  & 0.634  & 72.41  & 0.701  & 61.52  & 0.601  & 67.55 \textcolor{orange}{$\uparrow$}  & 0.709 \textcolor{red}{$\uparrow$}  & 72.38  & 0.701 & 65.27 \textcolor{orange}{$\uparrow$} & 0.621 \textcolor{blue}{$\uparrow$}\\
& MHA2       & 69.01  & 0.681  & 72.01  & 0.729  & 66.72  & 0.645  & 71.29 \textcolor{blue}{$\uparrow$}  & 0.732 \textcolor{orange}{$\uparrow$} & 74.56 \textcolor{blue}{$\uparrow$}  & 0.742 \textcolor{blue}{$\uparrow$} & 67.29 \textcolor{blue}{$\uparrow$} & 0.651 \textcolor{blue}{$\uparrow$}\\ 
\toprule
\end{tabular}
\end{table*}

\section{Experiments}
\subsection{Dataset Description}\label{AA}
For textual tasks, experiments are conducted on the benchmark datasets, laptop, restaurant from SemEval-2014\cite{b10} and twitter \cite{b11}. All aspect terms are labeled in three categories of polarities: positive, neutral, and negative. For multimodal (text-audio) ABSA task, this paper adds audio information to the above text level SemEval dataset by XunFei tools\footnote{https://www.xfyun.cn/services/online\_tts}. Each data segment contains aspect term, sentence and polarity. 

\subsection{Our Models and Comparison Models}
PosCE is a novel embeddings method for language understanding. Thus, this paper compares the models with or without PosCE to illustrate the effectiveness of PosCE. All baselines and their introduction are as follows:
\begin{itemize}[topsep=1pt]
\setlength{\itemsep}{2pt}
\setlength{\parsep}{0pt}
\setlength{\parskip}{0pt}
\item \textbf{SVM} \cite{b12} is a feature-based support vector machine.
\item \textbf{LSTMs} \cite{b13,b14} ATAE-LSTM adds the aspect embeddings to each word embeddings, which attention to get the final representation for classification.
\item \textbf{AOA} \cite{b15} ombines aspects and sentences jointly by Attention, which can capture the interaction.
\item \textbf{TNet} \cite{b16}, which is an aspect-based transformer model to gain bi-directional feature for classification.
\item \textbf{BERT-SPC} \cite{b17} sequential \emph{"[CLS]+context+[SEP]+ target+[SEP]"} is fed into the BERT.
\end{itemize}

And for text-audio ABSA task, some baselines are: 
\begin{itemize}[topsep=1pt]
\setlength{\itemsep}{2pt}
\setlength{\parsep}{0pt}
\setlength{\parskip}{0pt}
\item \textbf{MDRE} \cite{b18} has used RNNs to encode inputs text-audio level data followed by a DNN for classification.
\item \textbf{MCNN-LSTM} \cite{b19} has used a DCNN followed by RNNs to encode modalities followed by joint fusion.
\item \textbf{MHA2} \cite{b20} has used bi-directional recurrent encoders followed by a multimodality attention mechanism.
\end{itemize}

\subsection{Experimental Settings}
We download a \emph{Glove} \cite{b21} with dimension $k$=300 and \emph{bert-base-uncased} \cite{b22} with dimension $k$=768 and $L$=12 layers as pre-trained embedding space. The way to obatin the sentence embeddings is Mean Pooling, which has proved to have best effect \cite{b23}. Adam is used to optimize parameters with batch size of 16 and maximum epochs of 20. We set learning rates equal to $10^{-3}$ and $10^{-5}$ for RNN and Transformer based models respectively. L2 regularization is set to $10^{-5}$.

\subsection{Results}
This paper follow previous works and use Accuracy and F1 score to evaluate the performance of PosCE. For textual and multimodal ABSA, comparison results are reported in Table. Statistics demonstrate that the models use PosCE (right subtable) has a significant improvement. But the improvement of the text task is obviously higher than that of the text-audio task. Because PosCE is just used for language rather than both text and audio. For text-level ABSA task, there is 2.38\% average accuracy improvement after adding PosCE, 3.1\%, 1.16\% and 3.39\% improvement for laptop, restaurant and twitter respectively. It can be found that the performance of the dataset with higher overall accuracy is less improved. For multimodal-level ABSA task, 1.25\% average accuracy improvement is less than text only dataset obviously. The performance of using PosCE on the transformer based model is better than that on RNN, because RNN is serially processing tokens leads to the dispersion of the effect.

At the same time, taking LSTM model as an example, we explore when to update PosCE is the best time during training. Updating PosCE in epoch $=$1, $\leq$5, $=$5, $\leq$10, $=$10, $\leq$20, $=$20. The results are shown in Table 2. It can be seen that the best effect show up when epoch = 5 or epoch = 10. From the description of section 2, the updating of PosCE depends on the training process of the models. Therefore, the above experimental results show that: (1) PosCE should be updated after the model has a rough performance; (2) the results of updating PosCE repeatedly not only waste computing resources, but also can’t achieve good results; (3) The performance of existing models can be significantly improved by using PosCE. 


\begin{table}[!t]
\footnotesize
\caption{Experiments results for updating PosCE strategy.} 
\centering
\renewcommand\arraystretch{0.8}
\begin{tabular}{lccccccc}
\toprule
Epoch   & $= 1$   & $\leq 5$ & $= 5$            & $\leq 10$ & $= 10$           & $\leq 20$ & $= 20$  \\
\midrule
Acc & -0.6\% & 1.9\%  & \textbf{2.4\%} & 2.0\%   & 2.3\%          & 1.6\%   & -1.0\% \\
F1  & -0.017 & 0.021  & 0.034          & 0.027   & \textbf{0.036} & 0.023   & -0.004\\
\bottomrule
\end{tabular}
\end{table}

\section{conclusion}
This paper propose a novel Position-Guided Contributive Embeddings to capture the regulation from surrounding contexts to aspect term. It solves the "aspect with low-sensitive sentiment" issue in the ABSA task. Experiments on three public datasets demonstrate that using PosCE can advance the extant ABSA models. In addition, the further experiments show that PosCE should be updated once in the early stage of training, which can save computing resources and perform well.

\section*{Acknowledgment}
This research was supported by the Natural Science Foundation of Shanghai, China (No.20ZR1460500), the Shanghai Municipal Science and Technology Major Project (2021SHZD \\ ZX0100), and the Fundamental Research Funds for the Central Universities.

\vspace{12pt}

\end{CJK}
\end{document}